\documentclass[runningheads]{llncs}

\usepackage{booktabs}  
\usepackage{array}
\usepackage{bm}
\usepackage{multirow}
\usepackage{amssymb}
\usepackage{graphicx}
\usepackage{amsmath}
\usepackage[pagebackref=false,breaklinks=true,letterpaper=true,colorlinks,bookmarks=false]{hyperref}
\usepackage{cite}
\usepackage{url}
\usepackage{bbding}

\begin{document}
\title{Pairwise Relation Learning for Semi-supervised Gland Segmentation}
%
%
	\author{Yutong Xie$^{1}$\thanks{Y. Xie and J. Zhang contributed equally to this work. The work was partially done while the co-first authors were visiting 
	The University of Adelaide.}, 
	Jianpeng Zhang$^{1*}$, 
	Zhibin Liao$^{2,3}$, 
	Chunhua Shen$^{2}$,\\
	Johan Verjans$^{2,3}$,
	Yong Xia\Envelope$^{1,4}$\\
	}
	
	\institute{$^1$ National Engineering Laboratory for Integrated Aero-Space-Ground-Ocean Big Data Application Technology, School of Computer Science and Engineering, Northwestern Polytechnical University, Xi’an 710072, China	
	\\yxia@nwpu.edu.cn 
	\\$^2$ 
	The University of Adelaide, Australia
	\\$^3$ South Australian Health and Medical Research Institute, Adelaide, Australia
	\\$^4$ Research \& Development Institute of Northwestern Polytechnical University in Shenzhen, Shenzhen 518057, China
    }

	
%
\maketitle     
\begin{abstract}
		Accurate and automated gland segmentation on histology tissue images is an essential but challenging task in the computer-aided diagnosis of adenocarcinoma. Despite their prevalence, deep learning models always require a myriad number of densely annotated training images, which are difficult to obtain due to extensive labor and associated expert costs related to histology image annotations. In this paper, we propose the pairwise relation-based semi-supervised (PRS$^2$) model for gland segmentation on histology images. This model consists of a segmentation network (S-Net) and a pairwise relation network (PR-Net). The S-Net is trained on labeled data for segmentation, and PR-Net is trained on both labeled and unlabeled data in an unsupervised way to enhance its image representation ability via exploiting the semantic consistency between each pair of images in the feature space. Since both networks share their encoders, the image representation ability learned by PR-Net can be transferred to S-Net to improve its segmentation performance. We also design the object-level Dice loss to address the issues caused by touching glands and combine it with other two loss functions for S-Net. We evaluated our model against five recent methods on the GlaS dataset and three recent methods on the CRAG dataset. Our results not only demonstrate the effectiveness of the proposed PR-Net and object-level Dice loss, but also indicate that our PRS$^2$ model achieves the state-of-the-art gland segmentation performance on both benchmarks. 
\end{abstract}

	\section{Introduction}
    Quantitative measurement of glands on histology tissue images is an effective means to assist pathologists in diagnosing the malignancy of adenocarcinoma~\cite{Adenocarcinomas}. Manual annotation of glands requires specialized knowledge and intense concentration, and is often time-consuming. Automated gland segmentation avoids many of these issues and provides pathologists an unprecedented ability to reliably characterise and quantify glands. Although being increasingly studied to improve its accuracy, efficiency and objectivity~\cite{MILD-Net, SADL, DSE}, this task remains challenging mainly due to (1) inadequate training data with pixel-wise dense annotations and (2) small gaps and adhesive edges between adjacent glands.
	
	Currently, most available gland segmentation methods are based on deep convolutional neural networks (DCNNs)~\cite{MILD-Net, SADL, DCAN, DSE, Mchannel, FullNet}. 
	Chen~$et$ $al.$~\cite{DCAN} presented a deep contour-aware network that harnesses multi-scale features to separate glands from the background and also employs the complementary information of contours to delineate each gland. 
	Qu et al.~\cite{FullNet} proposed a full resolution convolutional neural network to improve the gland localization and introduced a variance constrained cross-entropy loss to advance the shape similarity of glands. 
	Yan et al.~\cite{SADL} developed a shape-aware adversarial learning model for simultaneous gland segmentation and contour detection. Although superior to the performance of previous solutions, the performance of these DCNN-based gland segmentation methods depends heavily on a substantial number of training images with pixel-wise labels, which are difficult to obtain due to the tremendous efforts and costs tied to the dense annotations of histology images.

	To alleviate the burden of data annotation, semi-supervised segmentation models have been developed to jointly use labeled and unlabeled data for co-training~\cite{my_semi}. Recent semi-supervised learning (SSL) methods are usually based on consistency regularization~\cite{semi_nips_survey}. Specifically, unlabeled data are exploited according to the smoothness assumption that certain perturbations of an input should not significantly vary the prediction~\cite{semi_nips_survey, semi_nips_dection, semi_ipmi, semi_cvpr_dual,semi_cvpr_s4l}.
	Nevertheless, these methods only measure the consistency between different perturbations of an input image. 
	In fact, different images may contain the same kind of foreground objects (e.g., glands). The objects on two images may share consistent representations in the feature space as long as they have the same semantic label. We advocate that such pairwise consistency should be explored to establish an unsupervised way to learn generalized feature representation from unlabeled data.
	
	In this paper, we propose the pairwise relation-based semi-supervised (PRS$^2$) model for gland segmentation on histology tissue images. This model is composed of a supervised segmentation network (S-Net) and an unsupervised pairwise relation network (PR-Net). The PR-Net is trained to boost its ability to learn both semantic consistency and image representation via exploiting the semantic consistency between each pair of images in the feature space. Since the encoders of S-Net and PR-Net share parameters, the ability learned by PR-Net can be transferred to S-Net to improve its segmentation performance. 
	Meanwhile, we employ the object-level Dice loss to impose additional constraints on each individual gland, and thus addresses the issues caused by touching glands. 
	The object-level Dice was originally proposed in~~\cite{Challenge} as a performance metric, but not as a loss function. We transform it as a loss and combine this loss with the pixel-level cross-entropy loss and global-level Dice loss to form a multi-level loss for S-Net.
	We evaluate the proposed PRS$^2$ model on the GlaS Challenge dataset and CRAG dataset and achieve superior performance over several recently published gland segmentation models.
	
	The contributions include: (1) proposing the pairwise relation interaction to exploit the semantic consistency between each pair of images in the feature space, enabling the model to learn semantic consistency and image representation in an unsupervised way; (2) transforming the object-level Dice evaluation metric as a loss and employing it to address the issues caused by touching glands; and (3) constructing the PRS$^2$ model that achieves the state-of-the-art gland segmentation performance on two benchmarks.
	
	\section{Method}
\begin{figure}[!t]
	\begin{center}
		{\includegraphics[width=1\linewidth]{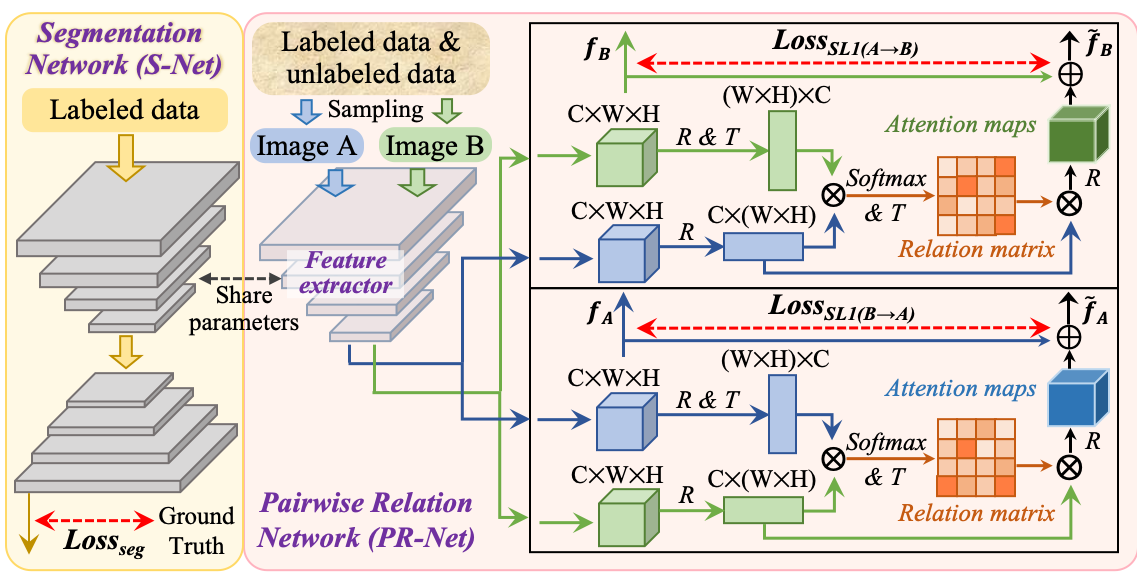}}
	\end{center}
	\vspace{-0.7cm}
	\caption{Diagram of the proposed PRS$^2$ model.}
	\label{fig:fig1}
	\vspace{-0.1cm}
    \end{figure}

	The proposed PRS$^2$ model has two major modules: the S-Net for supervised gland segmentation and PR-Net for unsupervised semantic relation learning (see Fig.~\ref{fig:fig1}). Let the labeled training set with $M$ images be denoted by $\bm{X}^L$, the unlabeled training set with $N$ images be denoted by $\bm{X}^U$, and the whole training image set be denoted by $\bm{X} = \bm{X}^L \cup \bm{X}^U$. The pipeline of this model can be summarized in two steps. First, the S-Net is trained on $\bm{X}^L$ for an initialization. Since the encoders of both networks share the same architecture and parameters, the encoders PR-Net is also initialized in this step. Then, both the S-Net and the PR-Net are jointly fine-tuned on $\bm{X}$ with the parameter-sharing mechanism.

    \vspace{+0.1cm}
    \noindent{\textbf{S-Net.}} 
    We use the DeepLabv3+ model~\cite{deeplabv3plus} pretrained on PASCAL VOC 2012 dataset~\cite{VOC} as S-Net. To adapt DeepLabv3+ to our task, we replace the last convolutional layer, which is task specific, with a convolutional layer that contains two output neurons to predict glands and background. The weights in this layer are randomly initialized, and the activation is set to the softmax function.
    
    We design the following multi-level segmentation loss $\mathcal{L}_{seg}$ for S-Net, defined as follows  
	\begin{equation}
	\mathcal{L}_{seg}=\mathcal{L}_{ce} + \mathcal{L}_{Dice} + \mathcal{L}_{objDice},
	\end{equation}
	where $\mathcal{L}_{ce}$ is the cross-entropy loss that optimizes pixel-level accuracy, $\mathcal{L}_{Dice}$ is the Dice loss that optimizes the overlap between the prediction and ground truth, and $\mathcal{L}_{objDice}$ is the object-level Dice loss. 
	Combining the first two losses is commonly used in many medical image segmentation applications and achieves remarkable success~\cite{CE_Dice_MICCAI18, CE_Dice_IJCAI19}. However, gland segmentation requires not only to segment the glands from background, but also to separate each individual gland from others. 
	The latter requirement is quite challenging due to the existence of touching glands. To address this challenge, we propose the object-level Dice loss as follows
	 \begin{equation}
    \mathcal{L}_{objDice}=\frac{1}{2}\left [ \sum_{i=1}^{n}\frac{|\bm{S}_{i}|}{\sum_{k=1}^{n}|\bm{S}_{k}|} \mathcal{L}_{Dice}(\bm{G}_{i}, \bm{S}_{i})+\sum_{j=1}^{m}\frac{|\tilde{\bm{G}}_{j}|}{\sum_{k=1}^{m}|\tilde{\bm{G}}_{k}|} \mathcal{L}_{Dice}(\tilde{\bm{G}}_{j}, \tilde{\bm{S}}_{j}) \right ],
	 \end{equation}
	 where $\bm{S}_{i}$ is the $i$th segmented gland, $\bm{G}_{i}$ is the ground truth gland that maximally overlaps $\bm{S}_{i}$, $\tilde{\bm{G}}_{j}$ is the $j$th ground truth gland, and $\tilde{\bm{S}}_{j}$ is the segmented gland that maximally overlaps $\tilde{\bm{G}}_{j}$. 
	 The $m$ and $n$ denote the total number of ground truth glands and segmented glands for an input image, respectively. In this definition, the first term measures how well each segmented gland overlaps its corresponding ground truth, whereas the second term measures how well each ground truth gland overlaps its corresponding segmented gland. This loss function considers the instance-level discrepancy between a segmentation result and its ground truth, and thus is able to help S-Net learn more discriminatory feature representations for gland segmentation.
    	
    \vspace{+0.1cm}
    \noindent{\textbf{PR-Net.}} 
    The PR-Net exploits the semantic consistency between each pair of images for unsupervised pairwise relation learning. It is a composition of three modules: (1) an image pair input layer, (2) an encoder $\mathcal{F}(\cdot )$ for feature extraction, and (3) a pairwise relation module (PRM). 
    The input layer accepts a pair of images ($\bm{x}_{A}$, $\bm{x}_{B}$), which are randomly sampled from the whole training set $\bm{X}$, as input.
    The encoder shares the identical architecture and parameters with the encoder of S-Net (i.e., modified aligned Xception), whose output can be formally presented as follows
    \begin{equation}
    \bm{f}_{A}=\mathcal{F}(\bm{x}_{A};\bm{\Theta})\in \mathbb{R}^{C\times H\times W}, \bm{f}_{B}=\mathcal{F}(\bm{x}_{B};\bm{\Theta})\in \mathbb{R}^{C\times H\times W},
	\end{equation}
	where $\bm{\Theta}$ denotes the parameters of the encoder, and $C, H$ and $W$ denote respectively the number of channels, height, and width of the encoded feature representation.

	The PRM is proposed to highlight the targets of the same semantic class but located on two images. To this end, we first calculate the consistency relation matrix $\mathbb{C}$ from $\bm{f}_{B}$ to $\bm{f}_{A}$ as follows
	\begin{equation}
	\mathbb{C}_{B\rightarrow A}= \mathrm{softmax} (R(\bm{f}_{A})^T \cdot R(\bm{f}_{B}))^T \in \mathbb{R}^{(H\times W)\times(H\times W)},
	\end{equation}
	where $R(\cdot)$ represents a reshape function which collapses the $H$ and $W$ dimensions into a single dimension with $H \times W$ elements, and the softmax function normalizes the elements in the second dimension. The ${\mathbb{C}_{(B\rightarrow A)}}_{i, j}$ measures the consistency $i$th flattened `pixel' (in the feature representation space) of $\bm{f}_{B}$ to $j$th `pixel' of $\bm{f}_{A}$, where a larger ${\mathbb{C}_{(B\rightarrow A)}}_{i, j}$ indicates a higher semantic consistency between these two `pixels'. 
	
	Next, we perform a matrix multiplication between $R(\bm{f}_{B})$ and $\mathbb{C}_{B\rightarrow A}$ to obtain the attention map $\mathbb{M}$ of $\bm{f}_{A}$, formulated as
	\begin{equation}
	\mathbb{M}_{A}= R^{-1}\big(R(\bm{f}_B) \cdot \mathbb{C}_{B\rightarrow A} \big) \in \mathbb{R}^{C\times H \times W},
	\end{equation}
	where $R^{-1}$ is a reverse operation of $R$, each element in $\mathbb{M}_{A}$ can be considered as a weighted sum of $\bm{f}_B$ over all positions, where the weights are determined by $\mathbb{C}_{B\rightarrow A}$. Finally, we add $\mathbb{M}_{A}$ to the feature map $\bm{f}_A$ via an element-wise summation to obtain the target-highlighted feature maps $\bm{\tilde{f}}_A$, show as follows
	\begin{equation}
	\bm{\tilde{f}}_A=\mathbb{M}_{A}+\bm{f}_A,
	\end{equation}
	Similarly, the $\bm{\tilde{f}}_B$ can be calculated as
	\begin{equation}
	\bm{\tilde{f}}_B=\mathbb{M}_{B}+\bm{f}_B,
	\end{equation}	

\begin{figure}[!t]
		\begin{center}
			{\includegraphics[width=0.9\linewidth]{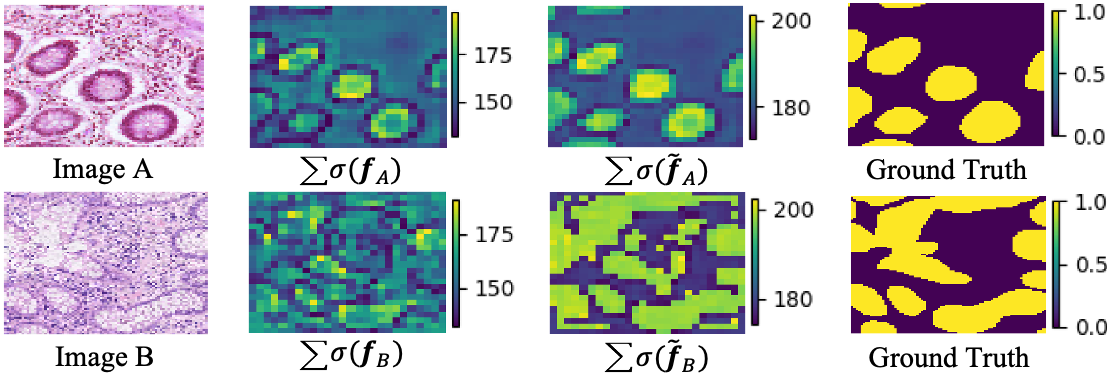}}
		\end{center}
		\vspace{-0.8cm}
		\caption{A pair of images and the corresponding channel-wise sum of $\sigma(\bm{f})$ and $\sigma(\bm{\tilde{f}})$
		}
		\label{fig:fig3}
		\vspace{-0.2cm}
	\end{figure}

	Both target-highlighted feature maps $\tilde{\bm{f}}_A$ and $\tilde{\bm{f}}_B$ have the merit of consistency relation information between $\bm{f}_A$ and $\bm{f}_B$, and thus can serve as the targets of PR-Net to enforce the model to increase the semantic consistency for any pair of image feature maps. Hence, the loss function of PR-Net can be expressed as
	\begin{equation}
	\mathcal{L}_{PR}=\underset{\mathcal{L}_{SL1}(B\rightarrow A)}{\underbrace{\mathcal{L}_{SL1}\left (\sigma(\bm{\tilde{f}}_{A}), \sigma(\bm{f}_{A}) \right )}}+\underset{\mathcal{L}_{SL1}(A\rightarrow B)}{\underbrace{\mathcal{L}_{SL1}\left (\sigma(\bm{\tilde{f}}_{B}), \sigma(\bm{f}_{B}) \right )}} ,
	\end{equation}
	where $\mathcal{L}_{SL1}$ is the smooth $L1$ loss, and $\sigma(\cdot)$ is the sigmoid function. Both $\bm{\tilde{f}}_A$ and $\bm{\tilde{f}}_B$, served as the target signals, do not perform back-propagation in each iteration. We also randomly select a pair of images and visualize their corresponding channel-wise sum of $\sigma(\bm{f})$ as well as $\sigma(\bm{\tilde{f}})$ in Fig.~\ref{fig:fig3} to show the superiority of $\bm{\tilde{f}}$.

	\vspace{+0.1cm}
    \noindent{\textbf{Optimization of PRS$^2$ model.}} 
	 The total loss of the proposed PRS$^2$ model is defined as the weighted sum of multi-level segmentation loss $\mathcal{L}_{seg}$ and unsupervised semantic consistency loss $\mathcal{L}_{PR}$ such that
	 \vspace{-0.2cm}
 	 \begin{equation}
	 \mathcal{L}_{total}=\mathcal{L}_{seg}+\alpha \mathcal{L}_{PR},
	 \vspace{-0.2cm}
	 \end{equation}
	 where $\alpha$ is a weighting factor that controls the contribution of unsupervised loss.
	 We adopt the Adam algorithm~\cite{Adam} with a batch size of 5 and 10 to train S-Net and PR-Net, respectively, and also set 20\% of the training set as a validation set to monitor the performance of both networks. The initial learning rate is set to 1e-4 in the initialization step and 5e-5 in the fine-tuning step. 

	\section{Experiments and Results}
	\noindent{\textbf{Materials.}} 
	We adopted the 2015 MICCAI Gland Segmentation (GlaS) challenge dataset~\cite{Challenge} and colorectal adenocarcinoma gland (CRAG) dataset~\cite{MILD-Net,CRAG} to evaluate the proposed PRS$^2$ model. 
	The GlaS dataset contains 85 training and 80 test images (60 in Part A; 20 in Part B). 
	The CRAG dataset has 173 training and 40 test images.
	When evaluating PRS$^2$ on the GlaS test set, the CRAG training set was considered as unlabeled training data, and vice versa.
	
	\noindent{\textbf{Evaluation Metrics.}} 
	On the GlaS dataset, three metrics officially suggested by the GlaS Challenge~\cite{Challenge} were calculated to assess the segmentation performance, including the object-level Dice (Obj-D) that represents the accuracy of delineating each individual gland, the object-level F1 score (Obj-F) that evaluates the accuracy of detecting each gland, and the object-level Hausdorff distance (Obj-H) that measures the shape similarity between each segmented gland and its ground truth. Meanwhile, all competing segmentation models were ranked according to each of these three metrics, and the sum of three ranking scores is calculated to measure the overall performance of each model. Note that a lower ranking score indicates better segmentation performance.
	
	\noindent{\textbf{Implementation Details.}} 
	In the training stage, we followed the suggestion in~\cite{DSE} to randomly crop patches from each training image as the input of both S-Net and PR-Net. The patch size was set to $416 \times 416$ on the GlaS dataset and $512 \times 512$ on the CRAG dataset. When training PRS$^2$ model, we resized CRAG patches to $416\times416$ if the labelled samples are from GlaS dataset, or resized GlaS patches to $512\times512$ if the labelled samples are from CRAG dataset.
	To further enlarge the training dataset, we employed the online data augmentation, which includes random rotation, shear, shift, zooming, and horizontal/vertical flip, and color normalization. 
	In the test stage, test time augmentations including cropping, horizontal/vertical flip and rotation, were also utilized to improve the robustness of segmentation. As a result, each segmentation result is the average of the results obtained on the original image and its three types of augmented copies. Moreover, the morphological opening using a square structure element with a size of $10 \times 10$ was finally performed to smooth segmentation results.
	
	\noindent{\textbf{Results on Two Datasets.}} 
	On the GlaS dataset, we compared the proposed PRS$^2$ model to five recently published gland segmentation models, including the deep contour-aware network (DCAN)~\cite{DCAN}, the minimal information loss dilated network (MILD-Net)~\cite{MILD-Net}, the shape-aware adversarial learning (SADL) model~\cite{SADL}, the rotation equivariant network (Rota-Net)~\cite{Rotanet}, the full resolution convolutional neural network (FullNet)~\cite{FullNet}, and the deep segmentation-emendation (DSE) model~\cite{DSE}. On the CRAG dataset, we compared our model to three models, i.e., DCAN, MILD-Net, and DSE. The performance of these models was given in Table~\ref{tab:tab1}. Note that the performance of all competing models was adopted in the literature, and the performance on the GlaS dataset is the average performance on test data part A and part B. Finally, we also compared our model to a typical semi-supervised (SS) method on both datasets, i.e., using a trained S-Net to generate segmentation predictions of unlabelled data and using a CRF-like approach to generate the proxy labels for fine-tuning the S-Net.
	
	It shows that our model achieves the highest Obj-D, second highest Obj-F, and lowest Obj-H on the GlaS dataset. Comparing to the DSE model that performs the second best, our model improves the Obj-D by 0.7\% and the Obj-H by 0.8. 
	On the CRAG dataset, it reveals that our model achieves the highest Obj-D, highest Obj-F, and lowest Obj-H, improving the Obj-D, Obj-F and Obj-H from 88.9\%, 83.5\% and 120.1, which were achieved by the second best model, to 89.2\%, 84.3\% and 113.1, respectively. 
	The results on both datasets indicate that the proposed PRS$^2$ model is able to produce more accurate for segmentation of glands and its performance is relatively robust.

	\begin{table}[!t]
	\caption{Gland segmentation performance of the proposed PRS$^2$ model and recently published models on both GlaS and CRAG datasets. M and R denote metric value and ranking score, respectively. Note that the performance on the GlaS dataset is the average performance on test data part A and part B}
	\label{tab:tab1}
	\vspace{-0.6cm}
	\begin{center}
    \begin{tabular}{c|c|c|c|c|c|c|c|c}
    \hline
    \multirow{2}{*}{Datasets}     & \multirow{2}{*}{Methods} & \multicolumn{2}{c|}{Obj-D} & \multicolumn{2}{c|}{Obj-F} & \multicolumn{2}{c|}{Obj-H} & \multirow{2}{*}{Rank sum} \\ \cline{3-8}
                                  &                          & M (\%)      & R     & M (\%)      & R     & M        & R        &                                                                         \\ \hline
    \multirow{8}{*}{GlaS dataset} & DCAN                     & 83.9            & 8        & 81.4            & 8        & 102.9        & 8           & 24                                                                      \\ \cline{2-9} 
                                  & MILD-Net                 & 87.5            & 6        & 87.9            & 5        & 73.7         & 6           & 17                                                                      \\ \cline{2-9} 
                                  & SADL                     & 87.3            & 7        & 88.9            & 3        & 76.7         & 7           & 17                                                                      \\ \cline{2-9} 
                                  & Rota-Net                 & 88.4            & 5        & 87.2            & 6        & 68.4         & 5           & 16                                                                      \\ \cline{2-9} 
                                  & FullNet                  & 88.5            & 4        & 88.9            & 3        & 63.0         & 4           & 11                                                                      \\ \cline{2-9} 
                                  & DSE                      & 89.9            & 2        & \textbf{89.4}            & \textbf{1}        & 55.9         & 2           & 5                                                                       \\ \cline{2-9} 
                                  & SS                       & 89.6            & 3        & 86.9            & 7        & 62.8         & 3           & 13                                                                      \\ \cline{2-9} 
                                  & \textbf{Our PRS$^2$}                    & \textbf{90.6}            & \textbf{1}        & 89.0            & 2        & \textbf{55.1}         & \textbf{1}           & \textbf{4}                                                                       \\ \hline
    \multirow{5}{*}{CRAG dataset} & DCAN                     & 79.4            & 5        & 73.6            & 5        & 218.8        & 5           & 15                                                                      \\ \cline{2-9} 
                                  & MILD-Net                 & 87.5            & 4        & 82.5            & 3        & 160.1        & 4           & 11                                                                      \\ \cline{2-9} 
                                  & DSE                      & 88.9            & 2        & 83.5            & 2        & 120.1        & 2           & 6                                                                       \\ \cline{2-9} 
                                  & SS                       & 87.6            & 3        & 81.6            & 4        & 145.0        & 3           & 10                                                                      \\ \cline{2-9} 
                                  & \textbf{Our PRS$^2$}                     & \textbf{89.2}            & \textbf{1}        & \textbf{84.3}            & \textbf{1}        & \textbf{113.1}        & \textbf{1}           & \textbf{3}                                                                       \\ \hline
    \end{tabular}
	\end{center}
	\vspace{-0.6cm}
	\end{table}

	\section{Discusses}
\begin{figure}[!t]
		\begin{center}
			{\includegraphics[width=1.0\linewidth]{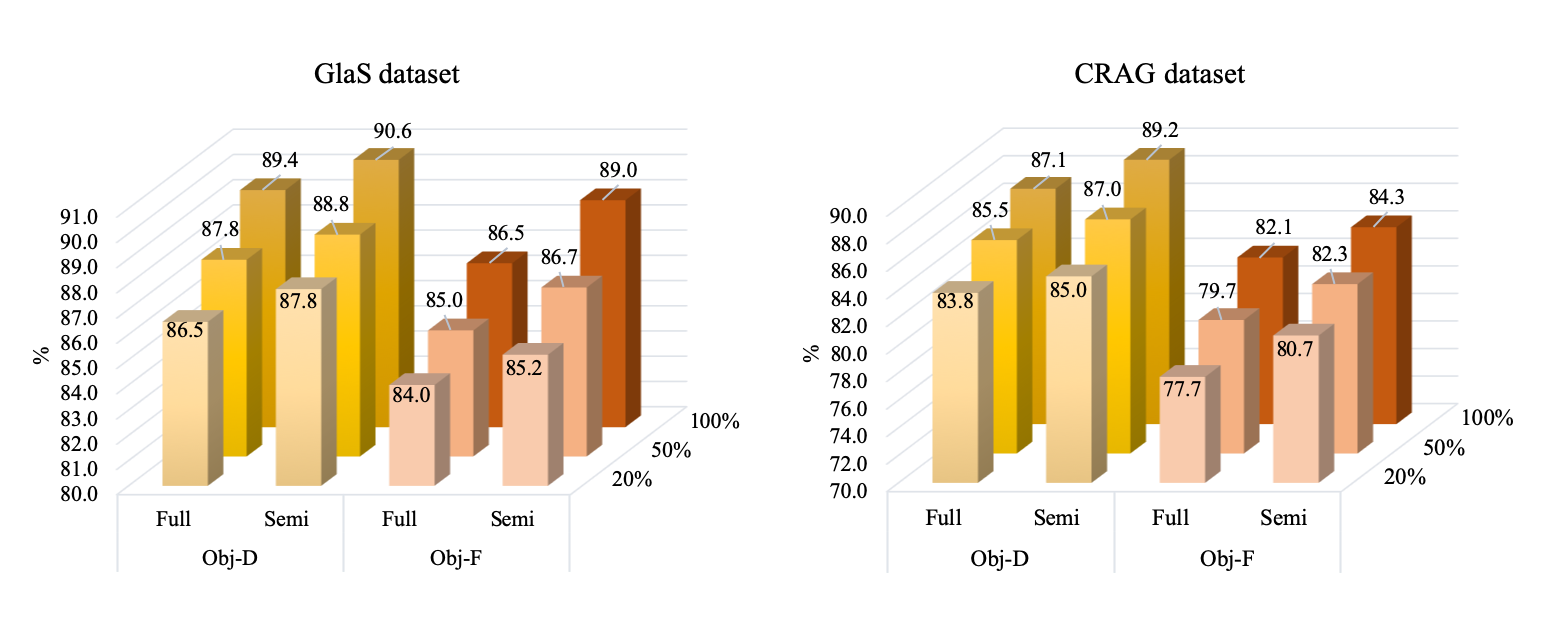}}
		\end{center}
		\vspace{-1.1cm}
		\caption{Obj-D and Obj-F values achieved on two datasets by our semi-supervised PRS$^2$ model and fully-supervised S-Net, when 20\%, 50\% and 100\% labeled training images}
		\label{fig:fig2}
		\vspace{+0.2cm}
	\end{figure}

	\noindent{\textbf{Trade-off between labeled and unlabeled data.}} A major advantage of our PRS$^2$ model is to use the unlabeled images to facilitate model training, leading to (1) less requirement of densely annotated training data or (2) improved segmentation performance when the labeled training dataset is small. To validate this, we kept the test set and unlabeled training set unchanged and randomly selected 20\% and 50\% labeled training images, respectively, to perform the segmentation experiments on both datasets again. As a control, we also used those selected labeled training images to train S-Net in a fully-supervised manner. The segmentation performance of our PRS$^2$ model and S-Net was shown in Fig.~\ref{fig:fig2}, from which three conclusions can be drawn. First, the segmentation performance of both models improves as the number of labeled training images increases. Second, using both labeled the unlabeled images, our model outperforms the fully-supervised S-Net steadily no matter how many labeled training images were used. More important, it is interesting that our model trained with 50\% labeled images can achieve a comparable performance with the fully-supervised S-Net trained with 100\% training data on both datasets. Similarly, it reveals that our model trained with 20\% labeled images can achieve a comparable performance with the S-Net trained with 50\% training data. It suggests that our model provides the possibility of using unlabeled data to replace almost half of labeled training images while maintaining the segmentation performance.
	
	\noindent{\textbf{Multi-level segmentation loss.}} 
	To demonstrate the performance gain resulted from the proposed multi-level segmentation loss, we also attempted to train the S-Net with different loss functions, including $\mathcal{L}_{Dice}$, $\mathcal{L}_{ce}$ and $\mathcal{L}_{Dice} + \mathcal{L}_{ce}$. The results in Table~\ref{tab:tab3} reveals that (1) using the combination of Dice and cross-entropy loss can produce higher Obj-F than using the Dice loss or cross-entropy loss alone, and (2) the superior performance of our multi-level loss over the combination of Dice and cross-entropy loss confirms the effectiveness of using the object-level Dice loss to pose constraints to each individual gland.
	
	\begin{table}[!t]
	\caption{Gland segmentation performance of S-Net obtained on two datasets when using different loss functions}
	\label{tab:tab3}
	\vspace{-0.9cm}
	\begin{center}
	\begin{tabular}{l|m{1.2cm}<{\centering}|m{1.2cm}<{\centering}|m{1.2cm}<{\centering}|m{1.2cm}<{\centering}|m{1.2cm}<{\centering}|m{1.2cm}<{\centering}}
		\hline
		\multirow{2}{*}{Loss functions} & \multicolumn{3}{c|}{GlaS dataset} & \multicolumn{3}{c}{CRAG dataset} \\ \cline{2-7} 
		& Obj-D     & Obj-F     & Obj-H     & Obj-D     & Obj-F     & Obj-H     \\ \hline
		$\mathcal{L}_{Dice}$                              & 86.5      & 86.2      & 75.3      & 84.7      & 78.9      & 174.9     \\ \hline
		$\mathcal{L}_{ce}$                             & 88.4      & 86.1      & 65.5      & 86.7      & 77.8      & 139.5     \\ \hline
		$\mathcal{L}_{ce} + \mathcal{L}_{Dice}$                             & 88.7      & 86.0      & 66.5      & 86.3      & 80.3      & 157.3     \\ \hline
		$\mathcal{L}_{ce} + \mathcal{L}_{Dice} + \mathcal{L}_{objDice}$(\textbf{Ours})                          &  \textbf{89.4}      &  \textbf{86.5}      &  \textbf{64.1}      &  \textbf{87.1}      &  \textbf{82.1}      &  \textbf{138.6}     \\ \hline
	\end{tabular}
	\end{center}
	\vspace{-0.5cm}
	\end{table}
	
    \noindent{\textbf{Complexity.}} Two parameter-sharing DCNNs in our PRS$^2$ model are trained using the open source Pytorch software packages. In our experiments, it took about 12 hours to train our PRS$^2$ model (2 hours for the initialization step and 10 hours for the fine-tuning step) and less than 1 second to segment each test image on a server with 4 NVIDIA GTX 2080 Ti GPUs and 128GB Memory.
	
	\section{Conclusion}
	In this paper, we propose the PRS$^2$ model for gland segmentation on histology tissue images, which consists of a supervised segmentation network with a newly designed loss and an unsupervised PR-Net that boosts its image representation ability via exploiting the semantic consistency between each pair of images in the feature space. Our results indicate that this model outperforms five recent methods on the GlaS dataset and three recent methods on the CRAG dataset. Our ablation study suggests the effectiveness of proposed loss and PR-Net. Although our model is built upon the specific application of gland segmentation, the pairwise relation-based semi-supervised strategy itself is generic and can potentially be applied to other deep model-based medical image segmentation tasks to reduce the requirement of densely annotated training images.

    \vspace{-0.2cm}
    \section*{Acknowledgment}
    \vspace{-0.2cm}
    Y Xie, J Zhang, and Y Xia were supported in part by the National Natural Science Foundation of China under Grants 61771397, in part by the Science and Technology Innovation Committee of Shenzhen Municipality, China, under Grants JCYJ20180306171334997, and in part by Innovation Foundation for Doctor Dissertation of Northwestern Polytechnical University under Grants CX202010.

	\vspace{-0.2cm}
	{
		\bibliographystyle{splncs04}
		\bibliography{egbib}
	}

\end{document}